\documentclass{article}
\usepackage{spconf,amsmath,graphicx,hyperref,amsfonts}
\usepackage{booktabs,multirow}
\usepackage{graphicx}
\usepackage{makecell}
\usepackage{bbm}
\usepackage{enumitem}

\usepackage[table]{xcolor} 
\definecolor{WinGreen}{RGB}{100,210,100}
\definecolor{RunGreen}{RGB}{200,235,200}
\newcommand{\win}[1]{\cellcolor{WinGreen}\textbf{#1}}
\newcommand{\ru}[1]{\cellcolor{RunGreen}#1}

\title{Cross Paraphrastic Invariance Learning for Hallucination Detection}
\name{
Shanshan Lin$^{1,\star}$ \quad Dongsheng Hong$^{1,\star}$ \quad Sibo Ju$^{1}$ \quad Chao Chen$^{2}$ \quad Sihong Xie$^{3}$ \quad Xiangwen Liao$^{1,\dagger}$
}

\address{
        $^{1}$Fuzhou University \qquad 
        $^{2}$Harbin Institute of Technology (Shenzhen)
        \\
        $^{3}$The Hong Kong University of Science and Technology (Guangzhou)
        \\
        $^{\star}$Equally contribution \qquad 
        $^{\dagger}$Corresponding author
        }

\begin{document}
\ninept
\maketitle

\begin{abstract}
Large language models (LLMs) frequently generate \emph{hallucinations}, which are unsupported by a source document. 
To avoid costly LLM-as-evaluator pipelines and the heavy annotation demands of existing classifiers, 
we propose \textbf{CPIL} (\emph{Cross Paraphrastic Invariance Learning}), a two-stage Siamese framework that maximizes the utility of existing labeled data. 
Concretely, CPIL constructs informative training \emph{pairs} by: (i) generating \textit{paraphrastic views} of each document-claim example as positives, and explicitly aligning their representations to enforce invariance to surface form; and (ii) mining \emph{same-document, opposite-label} pairs as hard negatives to sharpen document-sensitive decision boundaries. 
Then CPIL conduct a two-stage model training: 
Stage~1 performs contrastive pretraining to learn a paraphrase-invariant, grounding-aware embedding space; and Stage~2 attaches a lightweight classifier for binary groundedness. 
On the \textit{LLM-AggreFact} benchmark (11 tasks), CPIL surpasses strong baselines concerning F1 scores with only $\sim1\%$ labeled data, showing its \textit{prediction superiority} and \emph{label efficiency}.
%
\end{abstract}


\section{Introduction}
\label{sec:intro}
Large language models (LLMs) show impressive fluency and coherence across a wide range of tasks. 
However, a persistent challenge is \emph{hallucinations}, the generation of statements that are factually incorrect or unsupported by a given source document~\cite{rahman2025hallucination_review,maynez2020faithfulness}. 
Accurately detecting hallucination is essential for developing trustworthy, safety-aligned AI systems~\cite{krishna2025safety}.

\begin{figure}
    \centering
    \includegraphics[width=.95\linewidth]{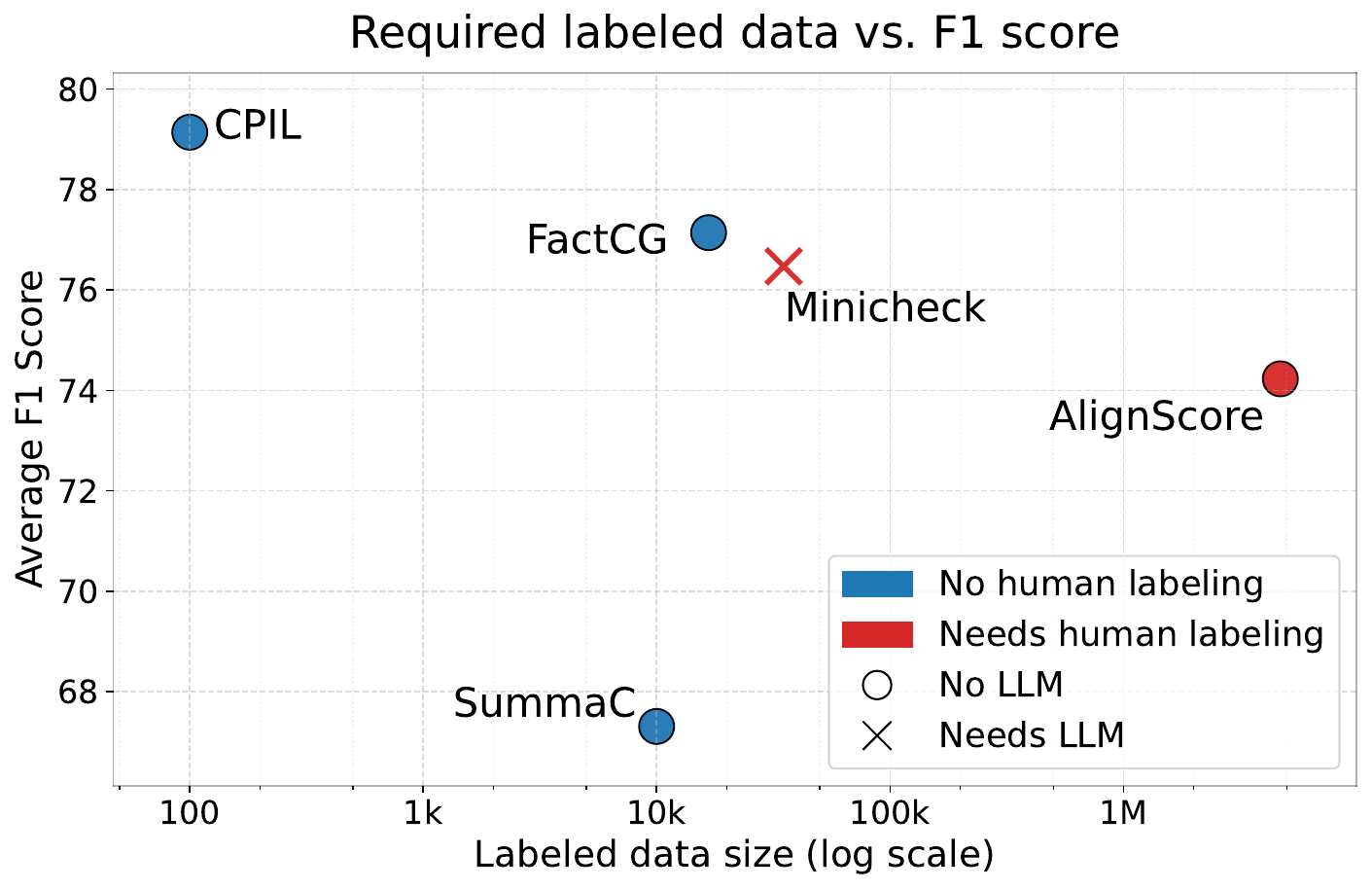}
    \caption{Required labeled data size \textit{vs.} model F1 score. 
    Colors and markers encode extra human and LLM labeling, respectively.
    }
    \label{fig:labeled_data_f1}
\end{figure}

Prior work on hallucination detection broadly falls into two families~\cite{rahman2025hallucination_review}. 
\textbf{(i)} \textbf{LLM-as-evaluator} methods (e.g., SelfCheckGPT~\cite{manakul2023selfcheckgpt}, SelfCheckAgent~\cite{muhammed2025selfcheckagent}, and G-Eval~\cite{liu2023geval}) query powerful LLMs to critique or verify generated content. 
However, they are often expensive at inference time and tokens, as a single instance may require multiple LLM prompts (e.g., iterative entity-level checks or chain-of-thought reasoning~\cite{wei2022cot}). 
\textbf{(ii)} \textbf{Factuality classifiers} train compact models to predict whether a claim is grounded in a source document,
and are typically much more efficient to deploy. 
For example, 
SummaC~\cite{laban2022summac} adapts pre-trained natural language inference (NLI) models for document-summary consistency; 
QAFactEval~\cite{fabbri2021qafacteval} probes documents by question answering; 
AlignScore~\cite{zha2023alignscore} aggregates NLI and factuality supervision; 
MiniCheck~\cite{tang2024minicheck} uses LLMs to label large collections of document-claim pairs; 
and FactCG~\cite{lei2025factcg} synthesizes multi-hop claims via graph-based augmentation. 
However, as illustrated in Fig.~\ref{fig:labeled_data_f1}, many high-performing detectors depend on large volumes of labeled or synthetic examples, which are costly to obtain, either through extensive human annotation of factual consistency or heavy use of LLMs for data generation and labeling~\cite{tang2024minicheck}. 

We propose \textbf{CPIL} (\emph{Cross Paraphrastic Invariance Learning}), a two-stage Siamese framework for document-grounded hallucination detection that \emph{better exploit existing labeled data} without additional human or LLM annotation. 
The key is to construct informative positive/negative training pairs from a few (e.g., 100) labeled samples and adopt a Siamese Network \cite{chen2021snx,he2018siamese} to leverage far more supervision signal than a standard classification approach. 
Concretely, we use back-translation~\cite{edunov2018understanding} to produce \textit{paraphrastic variants} of each document-claim instance. 
Instead of simply appending paraphrases to the training data \cite{kryscinski2019factcc}, 
CPIL explicitly aligns the representations of the same instance's \textit{cross-paraphrases} (as positives), encouraging invariance to lexical and syntactic variations. 
For negatives, rather than arbitrary cross-document mismatches~\cite{chen2020simclr}, CPIL samples \emph{same-document, opposite-label} pairs, yielding harder contrasts.  
After forming the pairing set tailored to hallucination detection, 
CPIL instantiates Siamese networks whose \textit{training} proceeds in two stages: 
\textit{Stage~1} performs contrastive pretraining to learn a paraphrase-invariant
embedding space by (i) pulling together cross-paraphrase views of the same instance and (ii) pushing apart pairs that share the document but differ in label.
\textit{Stage~2} attaches a lightweight classifier and optimizes a binary groundedness with limited labeled data. 

\begin{figure*}
    \centering
    \includegraphics[width=.9\linewidth]{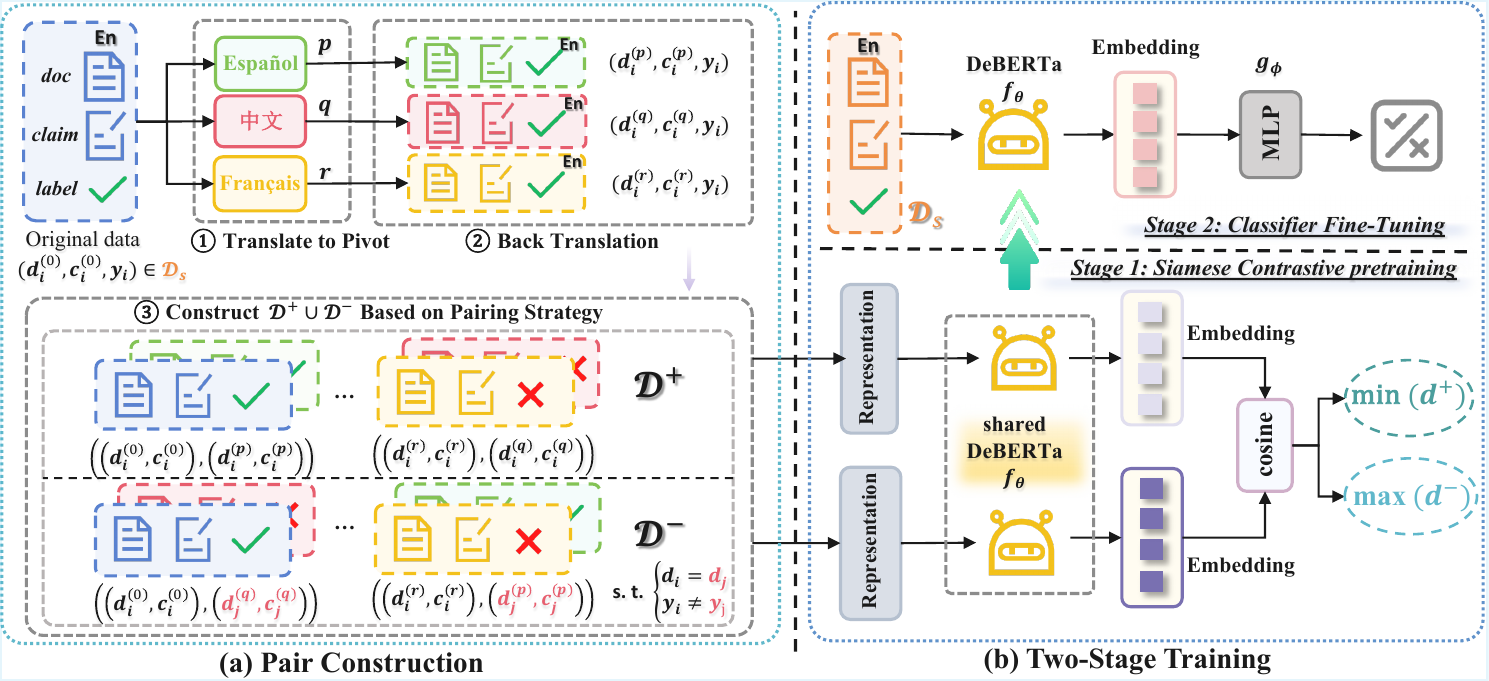}
    \caption{Framework of CPIL.
    \textbf{(a) Pair Construction}: generates label-preserving paraphrases
    (via back-translation), 
    and design a hallucination detection specified pairing strategy. 
    \textbf{(b) Two-Stage Training}: trains an encoder $f_\theta$ in Stage 1, and fine tune $f_\theta$ as well as a lightweight classifier $g_\phi$ in Stage 2. 
    }
    \vspace{-.5cm}
    \label{fig:framework}
\end{figure*}

Empirically, CPIL outperforms strong baselines, including FactCG~\cite{lei2025factcg}, MiniCheck~\cite{tang2024minicheck}, AlignScore~\cite{zha2023alignscore}, and SummaC~\cite{laban2022summac}, in average F1 score on \textbf{LLM-AggreFact} benchmark~\cite{tang2024minicheck}. 
Moreover, 
CPIL is markedly \emph{label-efficient}: it maintains strong performance with only a small fraction of supervised (e.g., $1\%$ of labels), surpassing baselines trained on the full (or larger) datasets. 
Ablation studies corroborate the importance of our pairing and training strategies.

In summary, our key contributions are:
(1) We design a pairing strategy tailored to hallucination detection: \textit{cross-paraphrase positives} 
and \textit{same-document, opposite-label hard negatives}.
(2) We propose CPIL, a two-stage Siamese detector that fully exploits existing labels via contrastive pretraining followed by lightweight classification, requiring no extra human or LLM supervision. 
(3) On LLM-AggreFact, CPIL attains superior average F1 with significantly reduced supervision, demonstrating strong label efficiency.

\section{Proposed Method}
\label{sec:method}

We begin by formalizing the task and notations in Sec.~\ref{sec:prelim}. 
We then introduce CPIL, our two-stage Siamese framework illustrated in Fig.~\ref{fig:framework}. 
Sec.~\ref{sec:views} describes how we obtain label-preserving paraphrastic views and construct informative training pairs. 
Sec.~\ref{sec:training} details contrastive pretraining and the classifier fine-tuning stages.

\subsection{Problem Setup}
\label{sec:prelim}
Document-grounded hallucination detection is framed as a binary classification task \cite{sansford2024grapheval}. 
Let training set be $\mathcal{D}=\{(d_i,c_i,y_i)\}_{i=1}^{N}$,
where \(d_i\) is a document, \(c_i\) is a claim about \(d_i\), and \(y_i\in\{0,1\}\) indicates whether the claim is \emph{supported} (\(1\)) or \emph{hallucinated} (\(0\)). The goal is to learn a detector
\(
h(d,c)=\Pr(y=1\mid d,c).
\)

Rather than collecting additional annotations~\cite{lei2025factcg}, CPIL maximizes the utility of the existing labeled subset \(\mathcal{D}_s\subset\mathcal{D}\) by constructing high-quality pairings. 
A naïve pairing strategy \cite{chen2020simclr} forms positives from two randomly sampled examples that share a label and negatives otherwise.
However, it may not suit to hallucination detection:
examples drawn from different sources can be topically and lexically dissimilar, yielding weak alignment signals. 
Thus, CPIL creates dense and informative supervision pairings by \emph{cross-paraphrase positives} and \emph{same-document hard negatives}.

\subsection{Paraphrastic Views and Pair Construction}
\label{sec:views}

\textbf{Label-Invariant Paraphrases.}
To encourage invariance to superficial rewording, we generate \emph{paraphrastic views} for each training example. 
Formally, given an original document-claim sample $(d_i,c_i,y_i) \in\mathcal{D}_s$, a paraphraser \(p\in\mathcal{P}\) produces a view \((d_i^{(p)},c_i^{(p)})\) that preserves meaning while altering wording and structure. 
Following the \emph{label-invariance} assumption \cite{bayer2023invariant}, any faithful paraphrase \((d_i^{(p)},c_i^{(p)})\) of \((d_i,c_i)\) inherits the same label 
$y_i$.

Among possible paraphrastic transformations (e.g. using an LLM to rewrite text or QA-style reformulation \cite{kryscinski2019factcc,yang2025metaqa}), 
we pick \textit{back-translation}, which translates a text from the source language to a pivot language and then back to the source language. 
Because it requires no human annotation or complex prompt engineering, and can be performed via low-cost off-the-shelf machine translation systems\footnote{We use Google Translation: \url{https://translate.google.com}.}~\cite{stahlberg2020translation_review}.
Notice that our pairing construction is agnostic to the specific paraphraser \cite{wei2019eda}, and the contribution lies in how these views are systematically exploited during pair construction.

\begin{table*}[h]
\centering
\caption{F1 score (in \%) across 11 tasks. Column best (without LLM) is \textbf{bold} with dark-green background, and runner-up is with light-green.}
\label{tab:main_result}
\resizebox{\textwidth}{!}{
\begin{tabular}{l|ccccccccccc|c}
\toprule
\multirow{2}{*}{Model}
& \multicolumn{2}{c}{AggreFact}
& \multicolumn{2}{c}{TofuEval}
& \multirow{2}{*}{WiCE}
& \multirow{2}{*}{REVEAL}
& \multicolumn{1}{c}{Claim}
& \multicolumn{1}{c}{Fact}
& \multicolumn{1}{c}{Expert}
& \multirow{2}{*}{LFQA}
& \multicolumn{1}{c|}{RAG}
& \multirow{2}{*}{AVG}
\\
& CNN & XSum & MediaS & MeetB &  &  & Verify & Check & QA &  & Truth &  \\
\midrule
SummaC-ZS     & 55.19 & 66.55 & 79.57 & 80.14 & 41.03 & 74.50 & 72.96 & 60.79 & 65.30 & 81.58 & 62.81 & 67.31 \\
SummaC-CV     & 66.41 & 48.56 & 69.70 & 77.87 & 50.98 & 70.77 & 68.36 & 57.53 & 65.21 & 79.16 & 56.92 & 64.68 \\
AlignScore    & 81.71 & 68.51 & 80.59 & 85.18 & 54.75 & 69.79 & 81.16 & 61.05 & 70.63 & 85.81 & 77.35 & 74.23 \\
MiniCheck     & 74.88 & \win{74.49} & 83.72 & 81.26 & 62.10 & 72.31 & \win{85.54} & 60.15 & \win{72.22} & \ru{88.15} & \ru{86.36} & 76.47 \\
FactCG        & 78.44 & \ru{74.46} & \ru{83.98} & 80.07 & \win{69.87} & \win{80.53} & 83.45 & \win{63.42} & 58.54 & \win{89.12} & \win{86.63} & 77.14 \\
\midrule
CPIL-1\%  & \win{91.99} & 73.47 & 83.02 & \ru{86.13} & \ru{69.20} & 77.13 & \ru{85.10} & \ru{61.46} & 70.57 & 87.85 & 84.74 & \win{79.14} \\
CPIL-2\%  & \ru{91.08} & 71.38 & \win{84.05} & \win{90.10} & 64.91 & \ru{77.77} & 81.74 & 58.36 & \ru{71.94} & \ru{88.90} & 84.79 & \ru{78.64} \\

\bottomrule
\end{tabular}
}
\end{table*}

\textbf{Cross-paraphrase Positive Pairs.}
Paraphrastic views provide natural \emph{positives} for contrastive training. 
Let \(\mathcal{V}=\{0\}\cup\mathcal{P}\) denote the set of available views, where view \(0\) is the identity \((d_i^{(0)},c_i^{(0)})=(d_i,c_i)\). 
We couple an instance with its own paraphrases (or two distinct paraphrases of the same instance) to form positive pairs. 
\begin{equation}
\mathcal{D}^{+}
=\Bigl\{\,\bigl((d_i^{(p)},c_i^{(p)}), \, (d_i^{(q)},c_i^{(q)})\bigr): \; 
p,q\in\mathcal{V}\Bigr\}.
\end{equation}
These cross-view positives encourage the encoder to map semantically equivalent inputs to nearby representations, promoting generalization to lexical and syntactic variation. 

\textbf{Same-document Negative Pairs.}
To drive fine-grained discrimination, we construct negatives from samples that \textit{share the same document} but have \textit{opposing labels}. 
Formally, 
\[
\Bigl\{\,\bigl((d_i,c_i),(d_j,c_j)\bigr)\;:\; 
d_i=d_j,\; 
y_i\neq y_j\,\Bigr\}.
\]
Holding the document \(d\) fixed enforces the model focus on fine-grained claim-evidence alignment within the shared document to distinguish supported from hallucinated assertions. 
%

To further increase difficulty and diversity, we optionally replace one or both sides with paraphrastic views while preserving the same-document constraint, yielding cross-view hard negatives: 
\begin{equation}
\mathcal{D}^{-}
=
\Bigl\{\, 
\bigl((d_i^{(p)},c_i^{(p)}),\,(d_j^{(q)},c_j^{(q)})\bigr):\; 
d_i = d_j, \;
y_i\neq y_j, \; p,q\in\mathcal{V}
\Bigr\}.
\end{equation}
The resulting hard negatives provide substantially stronger training signals than arbitrary cross-document mismatches.

\subsection{CPIL Network Training}
\label{sec:training}
CPIL comprises (i) an encoder \(f_{\theta}\) (e.g., DeBERTa~\cite{he2020deberta}) that maps a packed document-claim input \((d_i,c_i)\) to a single embedding $\mathbf{e}_i\in\mathbb{R}^{d}$, and (ii) a lightweight classifier \(g_{\phi}:\mathbb{R}^{d}\!\to\!\mathbb{R}\) that outputs a hallucination logit.


\textbf{Stage~1: Siamese Contrastive Pretraining.}
The encoder is trained to (i) pull together cross-paraphrase views of the same instances in $\mathcal{D}^+$ and (ii) repel pairs in $\mathcal{D}^-$ that share the same document but have opposite labels. 
Two weight-tied towers of \(f_{\theta}\) process paired inputs drawn from 
$\mathcal{D}^{+}\cup\mathcal{D}^{-}$.
%
Formally, 
we denote 
$
d^{+}=\cos\bigl(\mathbf{e}_i^{(p)},\,\mathbf{e}_i^{(q)}\bigr)
$
for a positive pair \(\bigl((d_i^{(p)},c_i^{(p)}), (d_i^{(q)},c_i^{(q)})\bigr)\in\mathcal{D}^{+}\),
and define $d^-$ similarly.
The objective is to minimize
\begin{equation}
\label{eq:ctr}
\mathcal{L}_{ctr}
= \mathbb{E}_{\mathcal{D}^+\,\cup\,\mathcal{D}^-}
\left[ \max\{0,\, d^{+}-d^{-}+\alpha\} \right],
\end{equation}
where \(\alpha>0\) is a margin, and we set it 1.5 in practice.
This yields an embedding space that is \emph{paraphrase-invariant} yet \emph{document-sensitive}.
One may alternatively employ other contrastive learning based loss, such as NT-Xent \cite{chen2020simclr} and SCL \cite{gunel2020supervised}.

\textbf{Stage~2: Classifier Fine-Tuning.}
We jointly optimize the pretrained encoder \(f_{\theta}\) and a newly attached lightweight head \(g_{\phi}\) (e.g., a single-layer MLP) with a small labeled set \(\mathcal{D}_s\):
\begin{equation}
\label{eq:stage2_loss}
\mathcal{L}_{ft} 
=\mathbb{E}_{(d,c,y)\sim \mathcal{D}_s}\!\left[
\operatorname{CE}\bigl(\, \sigma(g_{\phi}(f_{\theta}(d,c))), \, y\,\bigr)
\right],
\end{equation}
where \(\operatorname{CE}\) is the binary cross-entropy and $\sigma(\cdot)$ is the sigmoid function. 
We initialize \(f_{\theta}\) from Stage~1 and fine-tune it (or optionally freeze it).
Because Stage~1 has already shaped a decision-friendly representation, high accuracy is attained with \(|\mathcal{D}_s|\!\ll\!|\mathcal{D}|\).
At inference, prediction requires a single forward pass through \(f_{\theta}\) and \(g_{\phi}\).

\section{Experiments}

\subsection{Experiment Settings}
\label{sec:experiemnt_settings}

\textbf{Evaluation \& metrics.}
We evaluate on \textit{LLM-AggreFact}, an aggregate benchmark comprising 11 factuality datasets~\cite{tang2024minicheck}. 
We report per-task F1 (\%) and the macro-averaged F1 (AVG) across all tasks. 
We follow the official splits and evaluation protocols.

\textbf{Baselines.}
CPIL is compared against four powerful non-LLM detectors:
\textit{SummaC} (ZS and CV variants)~\cite{laban2022summac}, \textit{AlignScore}~\cite{zha2023alignscore}, \textit{MiniCheck}~\cite{tang2024minicheck}, and \textit{FactCG}~\cite{lei2025factcg}. 
When available, we use authors' released implementations and recommended checkpoints.

\textbf{Implementation details.}
For CPIL, the encoder \(f_\theta\) is instantiated with \textit{DeBERTa}~\cite{he2020deberta}, and the classifier \(g_\phi\) is a single-layer MLP. 
Paraphrastic views are generated via back-translation using three pivots, \(\mathcal{P}=\{\texttt{fr},\texttt{es},\texttt{zh}\}\) (French, Spanish, Chinese). 
Unless noted otherwise, optimization settings follow FactCG~\cite{lei2025factcg}, and we use a learning rate of \(1{\times}10^{-5}\).

\textbf{Data Budget and Pair Generation.}
Starting from the dataset \(\mathcal{D}\) released with FactCG~\cite{lei2025factcg} (approximately \(16.7\mathrm{k}\) labeled samples), 
our three-pivot paraphrasing and pairing procedure can yield up to more than \(350\mathrm{k}\) training pairs. 
To study label efficiency under capped supervision, 
$\mathcal{D}_s$ only uses $\sim2\%$ labeled data in $\mathcal{D}$, yielding about \(7\mathrm{k}\) training pairs; 
and a lower budget variant using about $\sim1\%$ labels leads to about \(4\mathrm{k}\) pairs.
We denote these configurations as \textit{CPIL-2\%} and \textit{CPIL-1\%}, respectively.

\subsection{Main Results}
\label{sec:main_results}

As reported in Table~\ref{tab:main_result},  CPIL attains the best macro average (79.14 with 1\% labels) and the second-best (78.64 with 2\% labels), outperforming all baselines on average. 
The largest margins are from \textit{CNN} and \textit{MeetB}, where paraphrase invariance and document-sensitive contrasts are especially beneficial given style variability and multi-speaker noise. 
Tasks such as WiCE and RAGTruth, which are sensitive to word sense and entity fidelity, remain challenging, while CPIL stays competitive.

Both the 1\% and 2\% configurations surpass all baselines trained at \textit{full} or \textit{larger} data scale. 
Thus, CPIL converts limited labels into richer supervision via pairwise alignment, yielding higher accuracy without additional annotation and inference overhead.

\begin{table*}[h]
\centering
\caption{\textbf{Ablation study}: CPIL's variants concerning different labeled data utilization strategy, and various pivots for back translation.} 
\label{tab:ablation_study}
\resizebox{\textwidth}{!}{
\begin{tabular}{>{\centering\arraybackslash}m{1.4cm}|c|ccccccccccc|c}
\toprule
\multirow{2}{*}{Group} 
& \multirow{2}{*}{Model}
& \multicolumn{2}{c}{AggreFact}
& \multicolumn{2}{c}{TofuEval}
& \multirow{2}{*}{WiCE}
& \multirow{2}{*}{REVEAL}
& \multicolumn{1}{c}{Claim}
& \multicolumn{1}{c}{Fact}
& \multicolumn{1}{c}{Expert}
& \multirow{2}{*}{LFQA}
& \multicolumn{1}{c|}{RAG}
& \multirow{2}{*}{AVG}
\\
& & CNN & XSum & MediaS & MeetB &  &  & Verify & Check & QA &  & Truth &  \\
\midrule
& CPIL-1\% & 91.99 & 73.47 & 83.02 & 86.13 & 69.20 & 77.13 & 85.10 & 61.46 & 70.57 & 87.85 & 84.74 & \textbf{79.14} \\
& CPIL-2\% & 91.08 & 71.38 & 84.05 & 90.10 & 64.91 & 77.77 & 81.74 & 58.36 & 71.94 & 88.90 & 84.79 & 78.64 \\
\midrule
\multirow[c]{3}{*}{\shortstack{Data\\Utilization}}
& ClsOnly  & 70.94 & 69.92 & 81.88 & 88.52 & 68.46 & 76.56 & 83.02 & 60.47 & 72.25 & 89.24 & 83.58 & 76.81 \\
& Append   & 78.30 & 69.20 & 82.46 & 86.74 & 69.90 & 77.77 & 86.41 & 61.28 & 65.15 & 87.47 & 86.10 & 77.34 \\
& RandPair & 77.16 & 74.15 & 79.84 & 83.52 & 70.31 & 76.49 & 85.04 & 61.69 & 70.70 & 87.69 & 85.33 & 77.44 \\
\midrule
\multirow[c]{3}{*}{\shortstack{Pivot\\Selection}}
& Pivot-\texttt{es} & 87.22 & 65.99 & 76.35 & 84.83 & 48.56 & 65.09 & 82.78 & 51.45 & 72.28 & 83.23 & 73.92 & 71.97 \\
& Pivot-\texttt{zh} & 84.41 & 70.57 & 84.34 & 81.07 & 66.91 & 75.77 & 85.22 & 58.82 & 71.04 & 88.41 & 84.54 & 77.37 \\
& Pivot-\texttt{fr} & 65.36 & 73.66 & 84.92 & 89.70 & 62.63 & 75.34 & 84.50 & 60.18 & 60.78 & 89.62 & 84.40 & 75.55 \\
\bottomrule
\end{tabular}
}
\end{table*}

\subsection{Sensitivity Analysis}
\label{sec:sensitivity_analysis}

\begin{figure}
    \centering
    \includegraphics[width=0.9\linewidth]{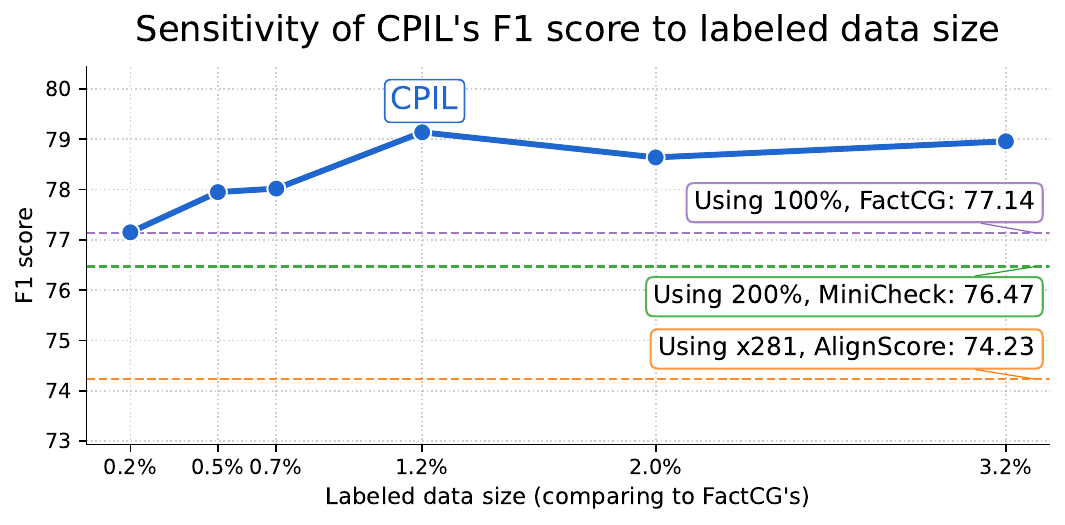}
    \caption{CPIL's F1 score when increasing fraction of labeled data used for pair construction, .e.g, $|\mathcal{D}_s| / |\mathcal{D}|$.}
    \label{fig:sensitivity}
\end{figure}

Fig.~\ref{fig:sensitivity} plots macro F1 as the fraction ($|\mathcal{D}_s| / |\mathcal{D}|$) of labeled examples used to form CPIL's pairs increases from 0.2\% to 3.2\% of the FactCG labels. 
Horizontal reference lines mark strong baselines, with their effective supervision levels annotated for context. 
Two observations stand out:

\textbf{(i) Strong low-budget performance.}
With only 0.2\% labels, CPIL already exceeds AlignScore and closely matches FactCG's full-data performance. 
As the labeled subset $\mathcal{D}_s$ grows (and thus $\mathcal{D}^{+}\!\cup\!\mathcal{D}^{-}$ enlarges), 
performance improves rapidly, peaking at 1.2\%.

\textbf{(ii) Mild plateau at moderate budgets.} 
Beyond 1.2\%, the curve stabilizes in a narrow band (about 78.6-79.0 F1). 
Modest non-monotonicity likely reflects 
(a) diminishing returns once the paraphrase-invariant geometry is well formed; 
(b) growing redundancy that increases the proportion of easy negatives and risks mild overfitting; 
and (c) finite encoder capacity. 
Practically, this indicates that strong accuracy is attainable without substantially scaling labels further, and additional gains may require increasing model capacity, or diversifying views-directions we leave to future work. 

Taken together, the curve highlights CPIL's practical superiority: high accuracy at low-percent label budgets, with clear avenues to trade additional compute or mining strategies for incremental gains.

\subsection{Ablation Study}
\label{sec:ablation}

We ablate CPIL along two axes: 
(i) data utilization (pairing/training strategy) and 
(ii) pivot selection 
(paraphraser). 
All ablations consume comparable supervision budgets by downsampling to
about $16\mathrm{k}$ training pairs/samples, 
and results are summarized in Table~\ref{tab:ablation_study}.

\textbf{(i) Data utilization.}
We consider three variants that use the same labeled pool \(\mathcal{D}\) but differ in how examples are consumed:
(a) \textit{ClsOnly}: omits contrastive pretraining (Stage~1), and trains a classifier on original $\mathcal{D}$ only.
(b) \textit{Append}: also skips Stage~1, but directly appends paraphrases to $\mathcal{D}$ without explicit view alignment. 
(c) \textit{RandPair}: forms positives/negatives by randomly pairing same-/opposite-label instances, irrespective of document identity.

\textbf{Pair construction matters.} 
Both Append and RandPair outperform ClsOnly, indicating that additional supervision signals,
whether via paraphrastic augmentation or pairwise signals, improves hallucination detection accuracy.
Notably, RandPair lags on MediaSum and MeetB, consistent with the intuition that random same-/different-label pairing induces many weak positives and easy negatives that fail to enforce document-sensitive distinctions. 
%
CPIL's superiority underscores the importance of the pairing/training strategies designed to hallucination detections.

\textbf{(ii) Pivot selection.}
We next examine single-pivot instantiations, denoted Pivot-\texttt{es}/\texttt{zh}/\texttt{fr}, which uses only one back-translation pivot (Spanish, Chinese, or French, respectively).
Single-pivot training consistently underperforms CPIL's multi-view setting. 
Among individual pivots, Pivot-\texttt{zh} is strongest overall, whereas Pivot-\texttt{es} lags markedly behind Pivot-\texttt{zh} and Pivot-\texttt{fr}. 

We hypothesize the contributing factor as:
Pivots that differ more in morphology and word order from \texttt{en} (e.g., \texttt{zh}) tend to induce richer syntactic rephrasings yet still return faithful \textit{en} renderings, 
which produce harder but valid positives. 
Conversely, closer pivots (e.g., \texttt{es}) often yield conservative paraphrases with limited structural variety, diminishing augmentation value. 

\textbf{Paraphrase diversity helps.}
Multi-pivot paraphrasing mitigates pivot-specific drawbacks: combining views increases syntactic/lexical diversity while averaging out pivot-induced errors. 
Consequently, CPIL’s multi-view version improves macro F1 over any single-pivot variant, indicating that \emph{view diversity under label invariance} could be the key.

\section{Conclusion}
\label{sec:conclusion}

We presented CPIL, a two-stage Siamese framework for hallucination detection that maximizes the utility of existing labels. 
CPIL converts individual labeled examples into \emph{pairs} by considering 
cross-paraphrase \emph{positives} and same-document, opposite-label \emph{hard negatives}. 
A contrastive pretraining stage learns a paraphrase invariant representation, followed by a lightweight classifier finetuning stage with single pass inference. 
On the \textit{LLM-AggreFact} benchmark, CPIL achieves state-of-the-art macro F1 with only $\sim$1\% of labels and remains competitive at even smaller budgets.

In the future, we will explore:
(i) improving faithfulness of back translation with safeguards against semantic drift, for example entailment based filters;
(ii) addressing the performance plateau at moderate label budgets using larger encoders (e.g., Flan-T5 \cite{chung2022flant5}), 
adaptive hard negative mining, 
and greater view diversity through more paraphrase strategies; and
(iii) extending CPIL to multi sentence claims, multi hop contexts, and multi modal grounding \cite{chen2024unified}. 


\section*{Acknowledgments}
This work was supported by the Natural Science Foundation of China (No. 62476060).
Chao Chen was also supported by the National Key Research and Development Program of China (No. 2023YFB3106504).
Sihong Xie was supported by the Department of Science and Technology of Guangdong Province (2023CX10X079), National Key R\&D Program of China (Grant No.2023YFF0725001), the Guangzhou-HKUST(GZ) Joint Funding Program (Grant No.2023A03J0008), and Education Bureau Guangzhou Municipality.

\bibliographystyle{IEEEbib}
\bibliography{main}

\end{document}